%% file: main.tex
\documentclass[10pt,twocolumn,letterpaper]{article}

\usepackage[pagenumbers]{iccv} 
\usepackage{multirow}
\usepackage{tabularx}
\usepackage{pifont}

\input{preamble}

\definecolor{iccvblue}{rgb}{0.21,0.49,0.74}
\usepackage[pagebackref,breaklinks,colorlinks,allcolors=iccvblue]{hyperref}

\title{LayerAnimate: Layer-level Control for Animation}

\author{%
  Yuxue Yang$^{1, 2}$~~~~~~~~Lue Fan$^{2}$~~~~~~~~Zuzeng Lin$^{3}$~~~~~~~~Feng Wang$^{4}$~~~~~~~~Zhaoxiang Zhang$^{1, 2}$\thanks{Corresponding author}\\
  $^1$School of Artificial Intelligence, University of Chinese Academy of Sciences\\
  $^2$NLPR \& MAIS, Institute of Automation, Chinese Academy of Science\\
  $^3$Tianjin University\hspace{4em}$^4$CreateAI\\
  \texttt{\{yangyuxue2023, lue.fan, zhaoxiang.zhang\}@ia.ac.cn}\\
  \texttt{linzuzeng@tju.edu.cn~~~ feng.wff@gmail.com}
  \vspace{-20pt}
}

\begin{document}
{
\makeatletter
\addtocounter{footnote}{1} 
\renewcommand\thefootnote{\@fnsymbol\c@footnote}
\makeatother
\maketitle
}
\input{sec/0_abstract}    
\input{sec/1_intro}
\input{sec/2_related}
\input{sec/3_method}
\input{sec/4_exp}
{
    \newpage
    \small
    \bibliographystyle{ieeenat_fullname}
    \bibliography{main}
}

\input{sec/X_suppl}

\end{document}

%% file: preamble.tex
\definecolor{blue}{RGB}{0,0,255}
\definecolor{orange}{RGB}{249, 169, 104}
\definecolor{yellow}{RGB}{245, 194, 66}
\definecolor{highlight}{RGB}{254, 254, 72}
\definecolor{gray1}{RGB}{128, 128, 128}
\definecolor{gray2}{RGB}{191, 191, 191}


%% file: sec/0_abstract.tex
\begin{abstract}
Traditional animation production decomposes visual elements into discrete layers to enable independent processing for sketching, refining, coloring, and in-betweening.
Existing anime generation video methods typically treat animation as a distinct data domain different from real-world videos, lacking fine-grained control at the layer level.
To bridge this gap, we introduce \textbf{LayerAnimate}, a novel video diffusion framework with layer-aware architecture that empowers the manipulation of layers through layer-level controls.
The development of a layer-aware framework faces a significant data scarcity challenge due to the commercial sensitivity of professional animation assets.
To address the limitation, we propose a data curation pipeline featuring Automated Element Segmentation and Motion-based Hierarchical Merging. Through quantitative and qualitative comparisons, and user study, we demonstrate that LayerAnimate outperforms current methods in terms of animation quality, control precision, and usability, making it an effective tool for both professional animators and amateur enthusiasts. This framework opens up new possibilities for layer-level animation applications and creative flexibility.
Our code is available at \url{https://layeranimate.github.io}.
\end{abstract}

%% file: sec/1_intro.tex
\section{Introduction}
\label{sec:intro}
Animation is a globally beloved art form, yet its production remains a complex process involving sketch drafting, refining, coloring, and in-betweening.
With the development of video generation models, automation technologies are increasingly being integrated into the animation production process.
Recent animation generation models~\cite{lvcd,tooncrafter,framer,anidoc} have adapted real-world generation models~\cite{svd,dynamicrafter} to achieve impressive results in interpolation and sketch coloring.
\begin{figure}
    \centering
    \includegraphics[width=\linewidth]{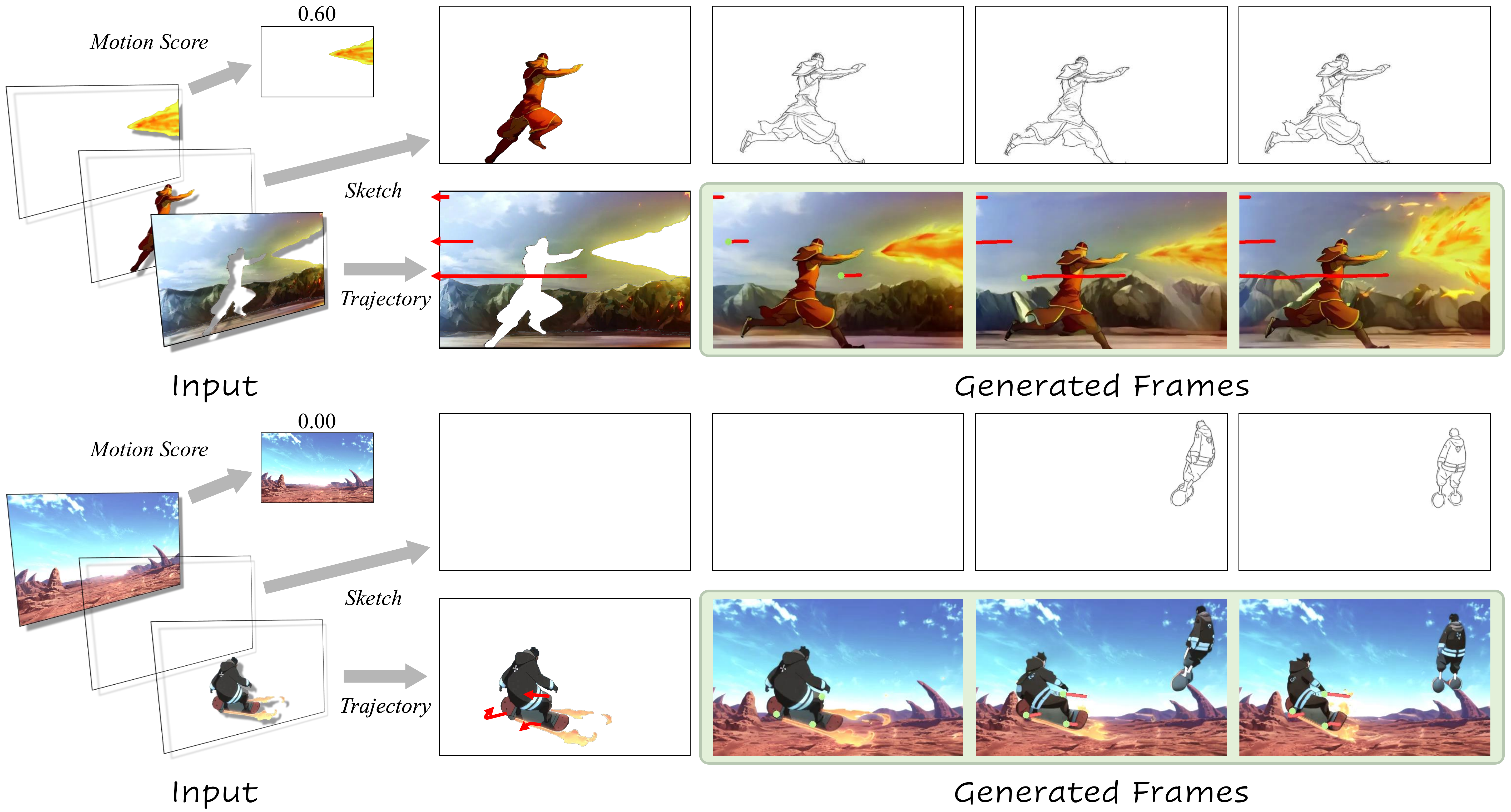}
    \vspace{-20pt}
    \caption{\textit{LayerAnimate} enables controllable video generation under multiple layer-level controls.}
    \label{fig:teaser}
    \vspace{-15pt}
\end{figure}
However, previous works typically treat animation as a distinct data domain compared to real-world videos, generating videos under frame-level controls.
They overlook a fundamental concept to animation, \textit{layer}, which allows independent controls on decomposed elements as depicted in \cref{fig:teaser}.
The principle of layer decomposition forms a foundational methodology across animation history, manifesting as stacked translucent celluloid overlays in classical hand-drawn production and evolving into digital layer hierarchies within modern software.
The hierarchical paradigm helps animators conduct nondestructive editing through layer isolation, enabling precise control over individual elements.

Considering the scarcity of layer data due to its commercial value, it is challenging to develop a video generation model supporting layer-level controls.
We design a layer curation pipeline comprising Automated Element Segmentation and Motion-based Hierarchical Merging to overcome the challenge.
We iteratively leverage SAM~\cite{sam1} and SAM2~\cite{sam2} for element segmentation, then merge over-segmented elements into layers based on their motion states with hierarchical clustering.

With the curated layer data, we propose \textit{LayerAnimate}, a framework that facilitates flexible composition of heterogeneous layer-level control signals and fine-grained manipulation of animation layers.
Initially, the frame-level reference image is decomposed into layer-level regions using layer masks from our curation pipeline, establishing explicit spatial correspondence between layer-level control signals and target regions.
Heterogeneous control modalities (e.g., motion scores, sketches, and trajectories) are injected into each layer through dedicated encoders. 
Following the encoding process, layer-level features are passed into ControlNet~\cite{controlnet} branches for independent processing, followed by cross-attention for feature fusion within the UNet.
LayerAnimate permits simultaneous manipulation of different elements under composite controls, which is unattainable in conventional frameworks.

We conduct extensive experiments and user studies across various video generation tasks under different conditions, i.e. first-frame Image-to-Video (I2V), I2V with trajectory, I2V with sketch, interpolation, interpolation with trajectory, interpolation with sketch, to demonstrate that  LayerAnimate is versatile and superior in terms of animation quality, control precision, and usability.
Our contributions are listed as follows.
\begin{itemize}
    \item We design a layer curation pipeline to automatically extract layer data from animations, addressing the challenge of limited layer data on the Internet.
    \item We propose a layer-level control framework, \textit{LayerAnimate}, that combines the traditional principle of layer decomposition with modern video generation models to achieve more precise animation control and generation.
    \item Extensive experimental results demonstrate the effectiveness and versatility of LayerAnimate on various tasks. It also supports innovative layer-level applications, such as a flexible composition of various layer-level controls.
\end{itemize}

%% file: sec/2_related.tex
\section{Related Works}
\paragraph{Video Diffusion Models.}
Video generation~\cite{lvdm, guo2024animatediff, sora,opensora,opensora_plan,easyanimate,cogvideox,pyramidal,moviegen,hunyuanvideo,ltx,allegro} has experienced significant advancements with the development of diffusion models~\cite{ho2020denoising,song2021denoising,dhariwal2021diffusion}.
Many methods~\cite{ho2022video, lvdm, ho2022imagen, singer2023makeavideo, xing2024make,guo2024animatediff} extend text-to-image diffusion architectures to generate temporally coherent videos.
However, it remains challenging to convey user intent exclusively through text.
To address this, several works~\cite{wang2024videocomposer, zhang2023i2vgen,svd,pixeldance,chen2023videocrafter1,seine,dynamicrafter,tooncrafter} incorporate images into diffusion models to enable video generation conditioned on given images.
To digest the reference image condition, a common approach used by VideoComposer~\cite{wang2024videocomposer}, VideoCrafter~\cite{chen2023videocrafter1}, and DynamiCrafter~\cite{dynamicrafter} is encoding the image through pre-trained CLIP or other well-designed image encoders before feeding it into diffusion models along with text prompts.
Furthermore, models like PixelDance~\cite{pixeldance}, SEINE~\cite{seine}, DynamiCrafter~\cite{dynamicrafter}, and ToonCrafter~\cite{tooncrafter} concatenate two different reference images with noisy frame latents to interpolate images with smooth transitions.
However, they fill the intermediate frames with placeholders, which underutilizes the conditions.
In contrast, our LayerAnimate assigns layers based on their motion states to the intermediate frame, allowing for injecting motion state.

\paragraph{Controllable Video Generation.}
Image-to-Video and interpolation models define videos' endpoints but struggle to provide motion information for intermediate frames.
Approaches~\cite{sparsectrl,tooncrafter,controlnext,lvcd,animateanyone,animatex,anidoc, draganything, tora} like SparseCtrl~\cite{sparsectrl} and ToonCrafter~\cite{tooncrafter} introduce an auxiliary branch for controllable video generation, inspired by ControlNet~\cite{controlnet}.
LVCD~\cite{lvcd} introduces a sketch-guided ControlNet to facilitate color transfer from the reference image to other frames.
However, these methods require frame-level control.
When applied to animation, such frame-level control will make regions without signals undergo unpredictable deformation.
The most recent work AniDoc~\cite{anidoc} facilitates high-quality animation with a reference character and sketch guidance without backgrounds, while it is tailored for characters.
Another classic control manner is movement control through trajectories, such as DragAnything~\cite{draganything} and Tora~\cite{tora}.
However, neither of them is adaptable to anime generation.
In this paper, our proposed \textit{LayerAnimate} allows users to provide layer-level control signals and supports applying multiple controls simultaneously in a more user-friendly manner for controllable anime generation. 

%% file: sec/3_method.tex
\section{Layer Curation}
\label{sec:layer_curation}
\begin{figure*}[ht]
    \centering
    \vspace{-15pt}
    \includegraphics[width=\linewidth]{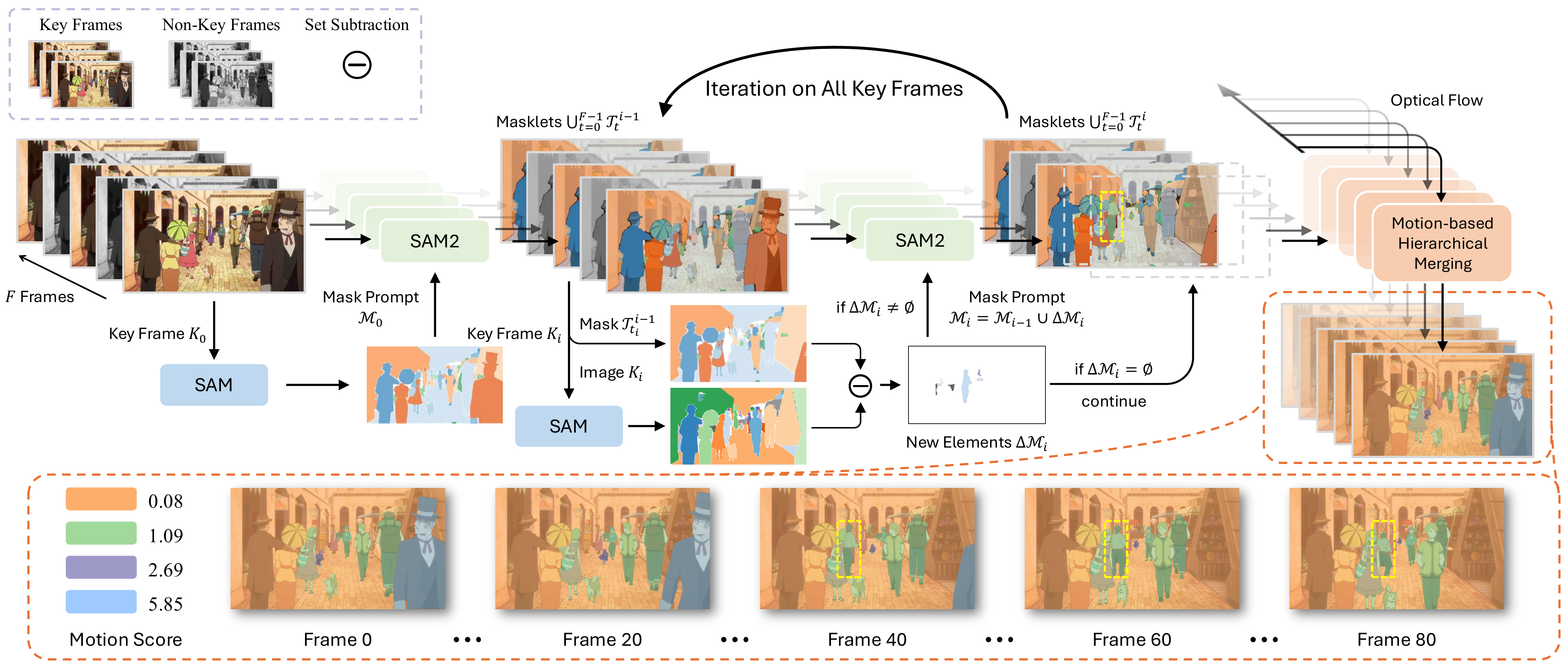}
    \caption{\textbf{Layer Curation Pipeline.} The bottom \textcolor{orange}{orange} dashed box illustrates curated layer masks with different motion scores, where motion scores remain temporally constant throughout the animation clip. \textcolor{yellow}{Yellow} dashed boxes denote new elements absent in the first frame, demonstrating our pipeline's capability to segment dynamically appearing elements. We transparently present some frames of masklets $\bigcup_{t=0}^{F-1}\mathcal{T}_t^i$ to highlight the new elements in Key Frame $K_i$.}
    \label{fig:data}
    \vspace{-12pt}
\end{figure*}
The construction of a well-curated dataset with detailed layer information is a prerequisite for training a layer-level controllable animation generation framework, which remains constrained by two critical challenges.
First, the commercial sensitivity of professional animation assets and the high cost of manual annotation make layer data hard to be scalable.
Second, unlike real-world video processing where depth estimation facilitates element stratification (e.g., MIMO~\cite{mimo}), the inherent 2D property of animations constrains reliable geometric cues for decomposing a frame into layers.
Conventional segmentation models applied to animations typically yield over-segmented color patches that lack semantics and are impractical for manipulation.
To address the challenges, we devise a novel layer curation pipeline comprising \textit{Automated Element Segmentation} and \textit{Motion-based Hierarchical Merging}, as illustrated in \cref{fig:data}.

\subsection{Automated Element Segmentation}
\label{sec:segmentation}
Taking advantage of recent advancements in visual foundation models~\cite{sam1, sam2}, we develop an iterative segmentation pipeline for automated element extraction in animation clips.
The process initiates with uniform temporal sampling at 4-frame intervals to establish Key Frames, where the first Key Frame $K_0$ is segmented via SAM~\cite{sam1} to generate atomic element masks $\mathcal{M}_0$.
These masks then serve as prompts, which are propagated to all $F$ frames in a clip through SAM2~\cite{sam2}, establishing initial masklets with temporal coherence.

Considering the frequent occurrence of new elements appearing in subsequent frames, the initial masklets cannot segment these elements.
Thus, we implement an iterative refinement to solve the issue.
We first denote the initial masklets as  $\bigcup_{t=0}^{F-1}\mathcal{T}_t^{i=0}$, where $\mathcal{T}_t^i$ denotes the refined masks for the $t$-th frame at the $i$-th iteration.
We detect new elements for each Key Frame $K_i$ with its frame index $t_i$ through mask set subtraction:
\begin{equation}
\Delta\mathcal{M}_i = \text{SAM}(K_i) \setminus \mathcal{T}_{t_i}^{i-1}.
\end{equation}
Any new elements $\Delta\mathcal{M}_i\neq\emptyset$ will update mask prompts
\begin{equation}
\mathcal{M}_i=\mathcal{M}_{i-1}\cup\Delta\mathcal{M}_i,
\end{equation}
which is repropagated through SAM2 to obtain refined masklets $\bigcup_{t=0}^{F-1}\mathcal{T}_t^i$.
If there is no new element, the masklets at iteration $i$ remain the same as iteration $i-1$.
The pipeline's iterative refinement enables coherent element extraction across the temporal dimension, even for the dynamically appearing elements. 

\subsection{Motion-based Hierarchical Merging}
\label{sec:motion_merging}
While SAM2 is capable of managing automated element segmentation for animations, it will cause an issue of over-segmentation.
This issue arises when regions that should belong to the same layer are divided by inner boundaries, resulting in a large count of elements.
If we regard each element as a layer, such over-segmentation breaks semantic objects into granular yet meaningless subdivisions, but also diminishes usability.

To address this, we introduce \textit{Motion-based Hierarchical Merging (MHM)}, designed to merge over-segmented masklets based on their motion states.
It is inspired by animation workflow, where animators dynamically merge or separate layers according to their motion states.
Firstly, we employ Unimatch~\cite{unimatch} to estimate optical flow, computing a motion score for each masklet by averaging flow magnitudes across all pixels in the masklet.
Notably, we do not use the direction of flow to represent motion state since pixels may move in diverse directions within a layer, such as dispersing smoke.
MHM regards masklets as nodes and constructs a treemap using hierarchical clustering based on motion scores, merging layers with similar motion scores from the bottom up.
Considering the variability in layer numbers during production, we do not restrict a fixed number of layers.
Instead, we define the maximum layer capacity $N$, which is much less than the number of masklets, and a motion score merging threshold $\eta_s$.
Layers are merged from the bottom up until the count of layers falls below the capacity $N$ and the motion score difference exceeds the threshold $\eta_s$.
The motion score of the final merged layer is set by averaging the motion scores from the merged layers.
A simple illustration of Motion-based Hierarchical Merging can be found in Supplementary Material.
\begin{figure*}[t]
    \vspace{-10pt}
    \centering
    \includegraphics[width=\linewidth]{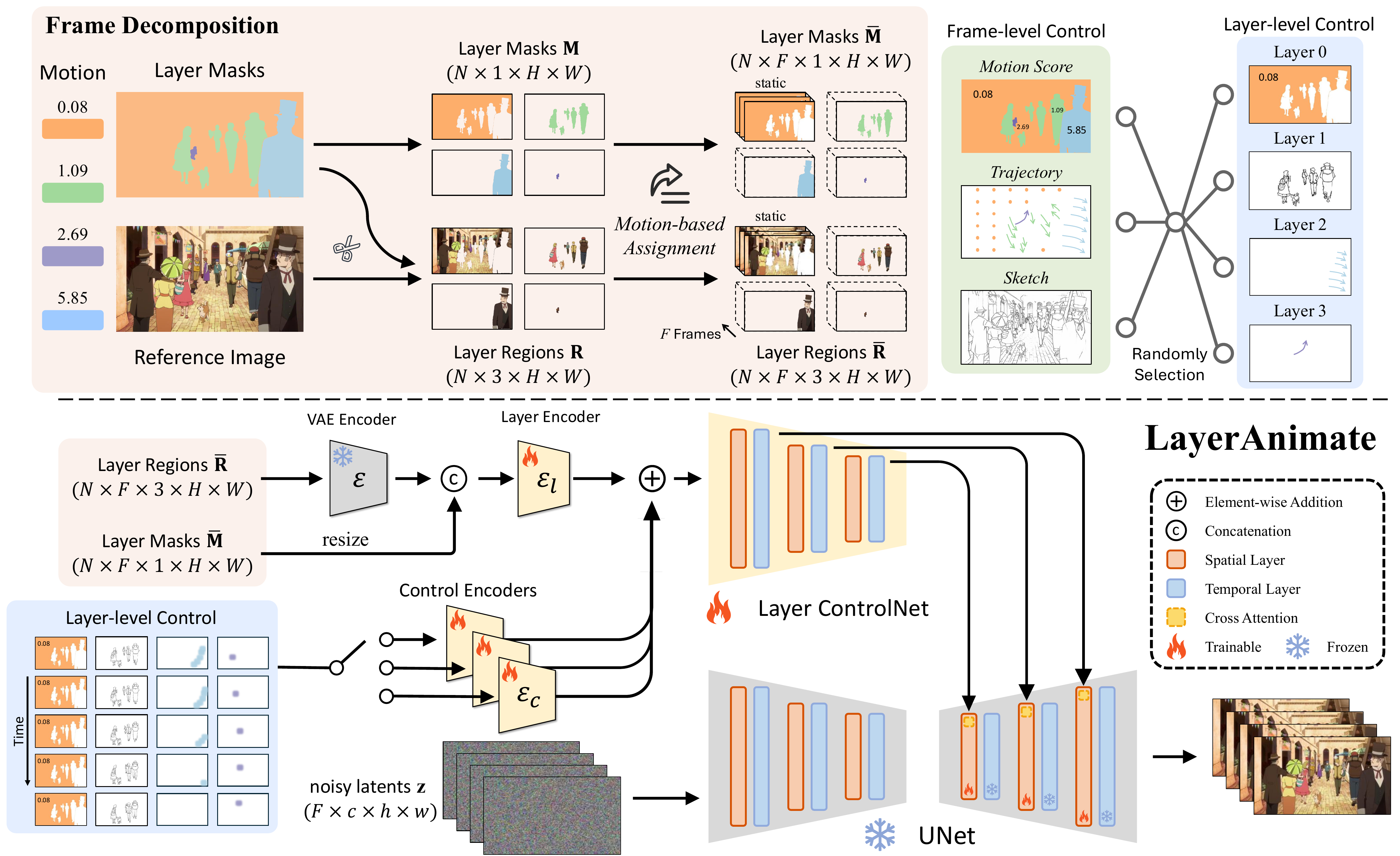}
    \caption{\textbf{Overview of \textit{LayerAnimate}}. LayerAnimate establishes a layer-level control architecture for animation generation. It enables the flexible composition of control signals at the layer level, allowing for injecting distinct conditions (e.g., motion scores, trajectories, and sketches) for different layers. For simplicity, the text and image injection branches are omitted from the core architecture schematic.}
    \vspace{-10pt}
    \label{fig:framework}
\end{figure*}

\section{LayerAnimate}
\label{sec:layer_framework}
Given a reference image $c_\text{image}$, layer masks $\mathbf M$, and layer-level control signals, our objective is to generate animation videos from Gaussian noise $\mathbf z$ through a conditional denoising network $\epsilon_\theta$.
Hence, we propose \textit{LayerAnimate}, a framework that enhances fine-grained control over layers within a video diffusion model, as illustrated in \cref{fig:framework}.

\subsection{Frame Decomposition}
\label{sec:frame_decomposition}
To unify the representation of layer information across various videos, we begin with padding the variable number of layer masks $\mathbf M$ to the fixed maximum capacity $N$.
We then decompose the reference image $c_\text{image}$ with binary layer masks $\mathbf M\in\mathbb R^{N\times 1\times H\times W}$ to get layer regions $\mathbf R\in\mathbb R^{N\times 3\times H\times W}$ and indicate layer motion state with their motion scores obtained from \cref{sec:motion_merging}.
With the layer information in hand, we need to consider it for non-reference frames across the temporal dimension.

Some multi-frame control methods, such as SparseCtrl~\cite{sparsectrl} and ToonCrafter~\cite{tooncrafter}, employ zero images to imply unconditional frames. Conversely, approaches like SVD~\cite{svd} and DynamiCrafter~\cite{dynamicrafter} that condition on a single reference image replicate the reference across all frames and then concatenate them with the input of the diffusion model.
In LayerAnimate, we integrate the aforementioned methods to propose \textit{Motion-based Assignment}.
It first categorizes layers into dynamic and static based on motion scores and a predefined threshold $\eta$, where the static layers are expected to remain unchanged along the temporal.
Specifically, we assign static layers from the reference to all $F-1$ non-reference frames, while assigning zero images to the $F-1$ non-reference frames of dynamic layers, where $F$ is the number of frames in the video.
Through the assignment, we unsqueeze layer masks and layer regions from $\mathbf M\in\mathbb R^{N\times 1\times H\times W}, \mathbf R\in\mathbb R^{N\times 3\times H\times W}$ to $\mathbf{\overline{M}}\in\mathbb R^{N\times F\times 1\times H\times W}, \mathbf{\overline{R}}\in\mathbb R^{N\times F\times 3\times H\times W}$ in the temporal dimension.

\subsection{Layer Controlling}
\label{sec:layer_control_preparation}
Following frame decomposition, precise control signal injection at the layer level is crucial for layer-level controllable animation generation.
Considering user accessibility, we implement three control modalities, which are in ascending order of control information: motion score (scalar fields), trajectory (directional guidance), and layer-level sketch (dense structural priors).
During training, layer-level control signals are randomly selected from frame-level signals through layer masks, enabling a flexible composition of control signals.
At inference, users can freely decompose the reference frame into layers and apply layer-level controls through an interactive interface.

\vspace{-13pt}
\paragraph{Motion Score.}
\label{sec:motion_information}
In Image-to-Video (I2V) task, motion is conventionally depicted by the text prompt; however, it's difficult for users to express precise motion descriptions for each layer.
Besides, certain elements like flames and particle effects, which are challenging to describe using trajectories, are common in animation clips.
Hence, we introduce layer-level motion scores to provide a more user-friendly control manner.
As detailed in \cref{sec:motion_merging}, we obtain layer motion scores $\mathbf s$ via optical flow estimation.
For consistent representation, we define an upper score $s_\text{max}$ and normalize $\mathbf s$ to $[0, 1]$ by $\mathbf s'=\lceil{\frac{\mathbf s}{s_\text{max}}}\rceil$.
The scores $\mathbf s'\in\mathbb R^{N\times 1}$ are spatially and temporally aligned with layer masks $\mathbf{\overline M}\in\mathbb R^{N\times F\times 1\times H\times W}$ through broadcasting and concatenated in the channel dimension.
Notably, layer masks $\mathbf{\overline M}$ only rely on the reference frame, eliminating the requirement for per-frame mask annotations.

\vspace{-13pt}
\paragraph{Trajectory.}
Trajectory offers enhanced spatial-temporal controllability compared to scalar motion scores.
We implement CoTracker3~\cite{cotracker3} to track the $60\times 60$ grid points across animation clips.
To filter out low-quality point trajectories wandering across different layers, we enforce constraints using masklets $\bigcup_{t=0}^{F-1}\mathcal{T}_t^i$ from \cref{sec:segmentation}.
We assign the masklets to each trajectory based on their coordinates in the first frame, then retain those trajectories maintaining more than 80\% overlap within the masklet to ensure layer-consistency.
The filtered trajectories are converted into a three-channel map, including one channel that indicates a Gaussian Heatmap like DragAnything~\cite{draganything} and the other two channels store a normalized offset map like Tora~\cite{tora}.
This hybrid representation combines the strengths of both forms: the heatmap channel resolves static/dynamic ambiguity in the offset map, i.e., static and uncontrolled regions are both zero, while the offset map models temporal correspondences between heatmap peaks in adjacent frames.
As demonstrated in \cref{sec:ablation}, the hybrid scheme achieves better performance.
\vspace{-13pt}
\paragraph{Sketch.}
Sketch enables precise manipulation of complex motions with dense structure guidance.
Unlike conventional complete frame-level sketch requirements, we randomly select layer-level sketches with curated layer masklets from \cref{sec:layer_curation} and remove the area of other layers to develop the capability of permitting partial sketching.

\subsection{Layer Feature Fusion}
As illustrated in \cref{fig:framework}, the decomposed layer regions $\mathbf{\overline{R}}$ are encoded into latent space by a VAE encoder.
To distinguish valid regions from invalid zero values, we resize layer masks $\mathbf{\overline{M}}$ to match the size of layer latents by bilinear interpolation for concatenation and further encode them with the layer encoder $\varepsilon_l$.
Layer-level controls are organized into the image format for subsequent encoding.
We implement conventional blocks to achieve an 8x spatial compression as control encoders $\varepsilon_c$ except for sketch, which is encoded by a VAE Encoder and a trainable convolution layer.

After the layer encoder and control encoders, the encoded layer features are combined and fed into ControlNet for parallel processing at the layer level, i.e., each layer is regarded as an independent sample.
Since the processed layer-level features $\mathbb R^{N\times F\times c\times h\times w}$ from ControlNet are $N$ times the number of frame-level features $\mathbb R^{F\times c\times h\times w}$ in the denoising UNet, we implement cross-attention to fuse layer features, where the frame-level features in UNet act as queries and the layer features serve as keys and values.
It's crucial to note that we introduce a validity mask to indicate padded layers, ensuring only valid layers participate in feature fusion.

\subsection{Training and Inference}
\paragraph{Training.}
During training, we optimize the conditional denoising network $\epsilon_\theta$, which consists of layer encoder $\varepsilon_l$, control encoders $\varepsilon_c$, UNet, and ControlNet.
The encoders $\varepsilon_l, \varepsilon_c$, the spatial layer in the decoder of UNet and ControlNet are trainable, while all other parameters are frozen.
The objective is given by:
\begin{equation}
\min\mathbb E_{\mathbf z_0,t,\epsilon\sim\mathcal N(0,\mathbf I)}\left[||\epsilon-\epsilon_\theta(\mathbf z_t;c, \mathbf{\overline{R}},\mathbf{\overline{M}}, \mathbf{L}_\text{c})||^2_2\right],
\end{equation}
where $\mathbf z_0$ represents the initial video latents from VAE encoder, $\mathbf z_t$ is the noised video latents at timestep $t$, $\mathbf{\overline{R}}, \mathbf{\overline{M}}$ denote the layer regions and masks obtained from \cref{sec:frame_decomposition}, $\mathbf{L}_\text{c}$ corresponds to layer-level controls, and $c$ indicates other conditions like the reference image $c_\text{image}$ and the text prompt $c_\text{text}$.
Moreover, we implement random control selection to enhance the model's robustness against diverse conditions.
We apply a 10\% dropout probability to layer masks simulating incomplete user annotations.
For each retained layer, the control among the three modalities will be randomly selected in the following probabilities: 20\% for motion score, 40\% for trajectory, and 40\% for sketch.
Since the three modalities are in ascending order of guidance information, the simultaneous application of weaker control does not provide additional guidance when selecting a strong control; therefore, we only select one control for each layer.
\vspace{-13pt}
\paragraph{Inference}
During inference, LayerAnimate allows users to generate layer masks on the reference image by simply clicking using SAM~\cite{sam1}.
Users can freely input distinct controls for different layers to generate an animation video tailored to the users' specifications.

%% file: sec/4_exp.tex
\section{Experiments}
\subsection{Implementation}
During the layer curation phase, we collect a considerable number of raw animation videos, which are systematically cleaned following OpenSora~\cite{opensora}.
On this basis, we curate layer data through our layer curation pipeline.
Throughout the process, we define the maximum layer capacity as $N=4$ and set the motion score merging threshold $\eta_s=1.0$.
The pipeline yields a dataset of 665K clips, ranging from 16 to 128 frames per clip, from which 1K clips are randomly selected as the evaluation set.

We adopt the pre-trained UNet from ToonCrafter~\cite{tooncrafter}, designed for cartoon interpolation, as our denoising UNet.
We replace its specially designed interpolation-oriented VAE with a standard VAE utilized in an I2V model DynamiCrafter~\cite{dynamicrafter} in the I2V task.
In LayerAnimate, the control encoders $\varepsilon_c$ and layer embedding $\varepsilon_l$ are implemented with convolutional blocks.
During the training, we classify layers with motion scores below $\eta=0.1$ as static and define the upper score $s_\text{max}=30.0$.
The sketches utilized in the experiments are extracted from original videos using the method in~\cite{sketch}.

All experiments are conducted over 30,000 steps using AdamW~\cite{adamw} optimizer with a learning rate of 2e-5 on 32 NVIDIA A100 GPUs.
The total batch size is set to 96.
Our LayerAnimate, with a maximum layer capacity of $N=4$, is trained to generate 16 frames at a resolution of $320\times 512$ on our collected anime dataset.
\subsection{Comparison}
\label{sec:comparison}
To demonstrate the versatility of our model, we conduct comparisons across six video generation tasks under different conditions: first-frame Image-to-Video (I2V), I2V with trajectory, I2V with sketch, interpolation, interpolation with trajectory, and interpolation with sketch.
For these tasks, we compare our method against the latest representative state-of-the-art methods: SEINE~\cite{seine}, DynamiCrafter~\cite{dynamicrafter}, and CogVideoX~\cite{cogvideox} for I2V and interpolation tasks, DragAnything~\cite{draganything} and Tora~\cite{tora} for I2V with trajectory task, Framer~\cite{framer} for interpolation with trajectory task, AniDoc~\cite{anidoc} and LVCD~\cite{lvcd} for the I2V with sketch task, and ToonCrafter~\cite{tooncrafter} for interpolation and interpolation with sketch tasks.
\vspace{-10pt}
\paragraph{Discussion.} Here we first briefly discuss the core differences between our methods and some related methods. (1) DragAnything~\cite{draganything} assigns trajectories to distinct entities based on masks, which is similar to our concept of layer-level control; however, its control is limited to the simple displacement of entities.
(2) AniDoc~\cite{anidoc} is tailored for character sketch coloring with the reference character specification.
Here, we take the first frame as the reference.
(3) Framer~\cite{framer} enables interpolation with given trajectories, where we provide the trajectories obtained by CoTracker3~\cite{cotracker3}.
To ensure a fair comparison, we do not input motion scores in I2V and interpolation tasks and adopt the same trajectories and sketches as the counterparts in related tasks.

\vspace{-12pt}\paragraph{Quantitative Comparison.}
\begin{figure*}[ht]
    \centering
    \vspace{-30pt}
    \includegraphics[width=\linewidth]{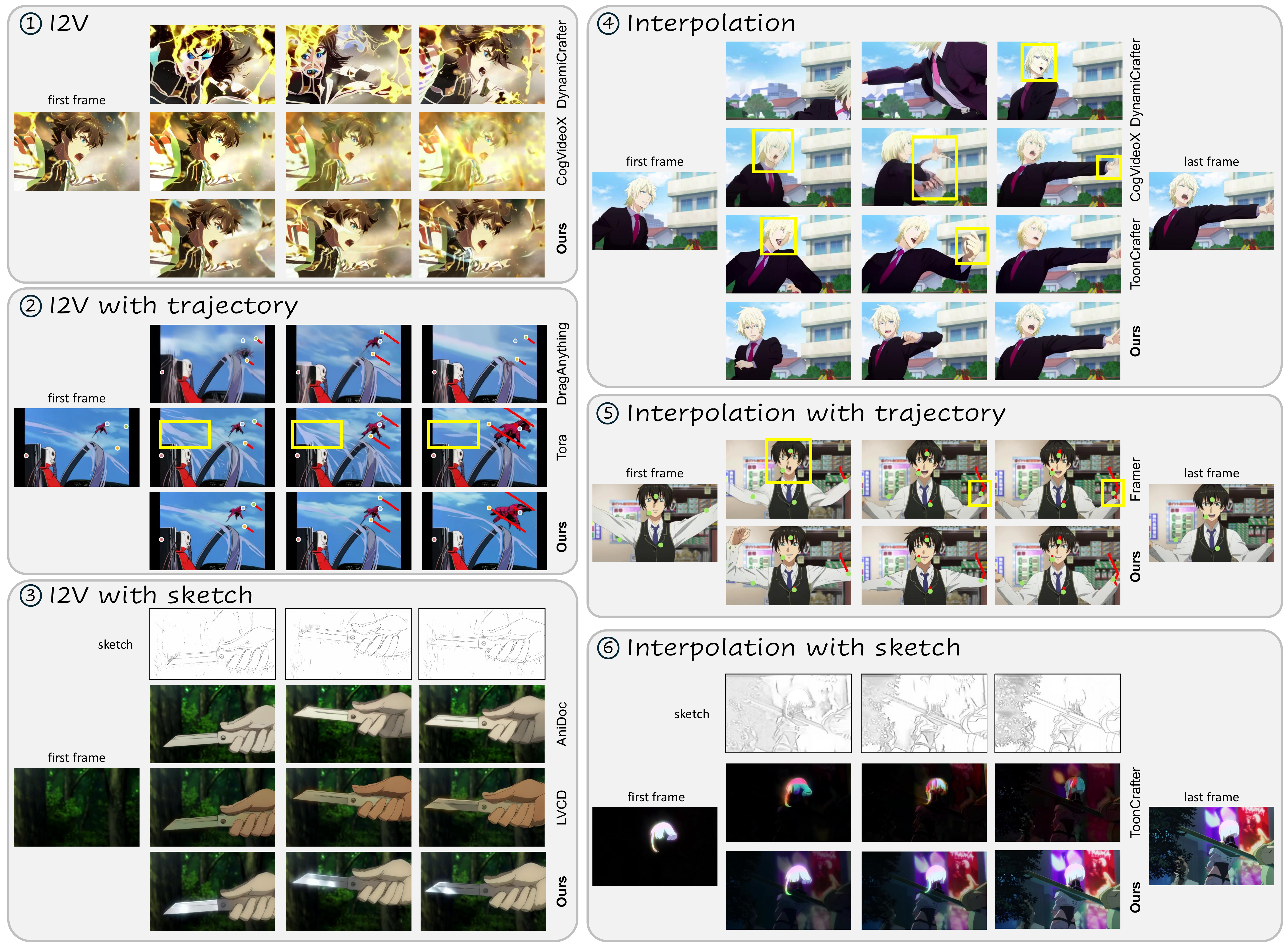}
    \caption{\textbf{Qualitative comparison with other competitors.} We select several clips to exemplify the representative characteristics of animation, including particle effects in \ding{172}, a knife appearing off-screen \ding{174}, and an unconventional fade-in visual style in \ding{177}. We provide the \textbf{corresponding videos} in the supplementary materials, offering more clear and vivid comparisons.}
    \vspace{-15pt}
    \label{fig:compare}
\end{figure*}
\input{table/compare}
To evaluate the quality of the generated videos in both spatial and temporal domains, we employ FVD~\cite{fvd} and FID~\cite{fid} metrics.
Additionally, to assess reconstruction quality in sketch-conditioned tasks, we adopt LPIPS~\cite{lpips}, PSNR, and SSIM to measure the similarity between the generated videos and the original videos.
As presented in \cref{tab:comparison}, our method demonstrates superior performance in all tasks.
Although we only demonstrate the performance with a certain single control in \cref{tab:comparison} for fairness, our approach also allows for the combination of multiple control modalities, indicating that our model possesses greater applicability, which can be seen in \cref{sec:layer_control}.

\vspace{-12pt}\paragraph{Qualitative Comparison.}
\label{sec:qualitative}
Unlike real-world videos, anime videos feature \textit{special effects}, \textit{objects appearing from nowhere}, and \textit{unconventional visual styles}.
We select several representative clips for qualitative comparison, as depicted in \cref{fig:compare}.
\begin{itemize}[leftmargin=*]
    \item For I2V, DynamiCrafter struggles to maintain character consistency, CogVideoX doesn't animate any elements but merely blurs the image, whereas our method not only generates particle effects but also preserves the character's facial consistency after particles pass across it.
    \item For I2V with trajectories, the flying red mecha controlled by DragAnything disappears halfway through, and the glass canopy of the aircraft not controlled by Tora fails to maintain consistency.
    Our method exhibits excellent tracking on the movement of the red flying mecha, and enables aircraft to be unchanged through a fixed point trajectory.
    \item For I2V with sketches, our method generates the knife with luster, exhibiting greater detail.
    \item For interpolation, our method generates more reasonable arm movements and facial expressions.
    \item For interpolation with trajectories, our method achieves excellent tracking and generates a more natural facial expression than Framer.
    \item For interpolation with sketches, which involves a fade-in scene, ToonCrafter fails to reveal the background properly, and the character's hair color alters over frames.
    Our method maintains consistent hair color while accurately generating the intended fade-in visual style.
\end{itemize}

\begin{figure}[t]
    \centering
    \includegraphics[width=\linewidth]{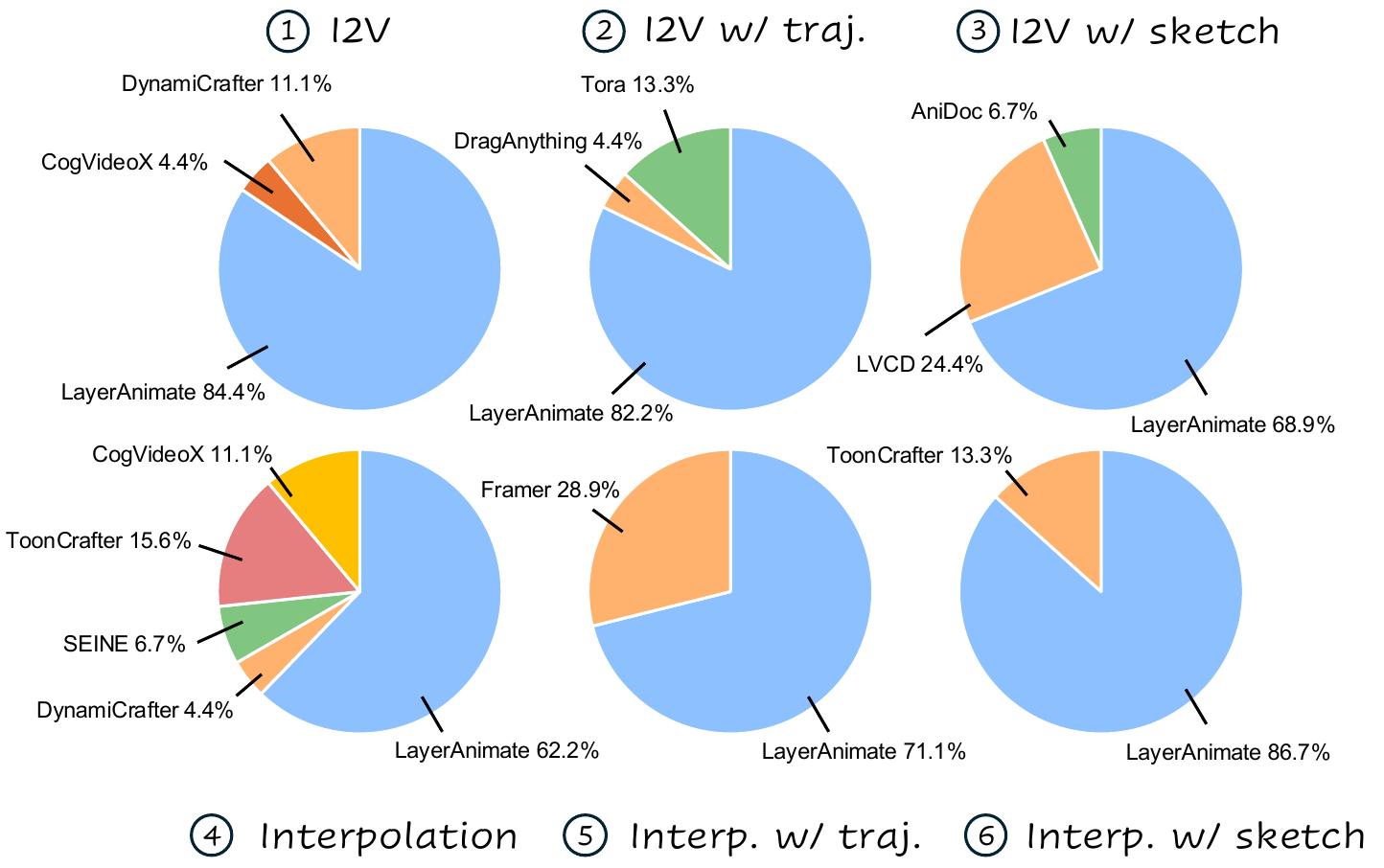}
    \caption{\textbf{Voting results of the user study.} LayerAnimate exhibits superior performance across different tasks. Interp.: Interpolation. traj.: trajectory.}
    \label{fig:user_study}
    \vspace{-17pt}
\end{figure}

\subsection{Composite Control}
\label{sec:layer_control}

\begin{figure*}[ht]
    \centering
    \vspace{-25pt}
    \includegraphics[width=\linewidth]{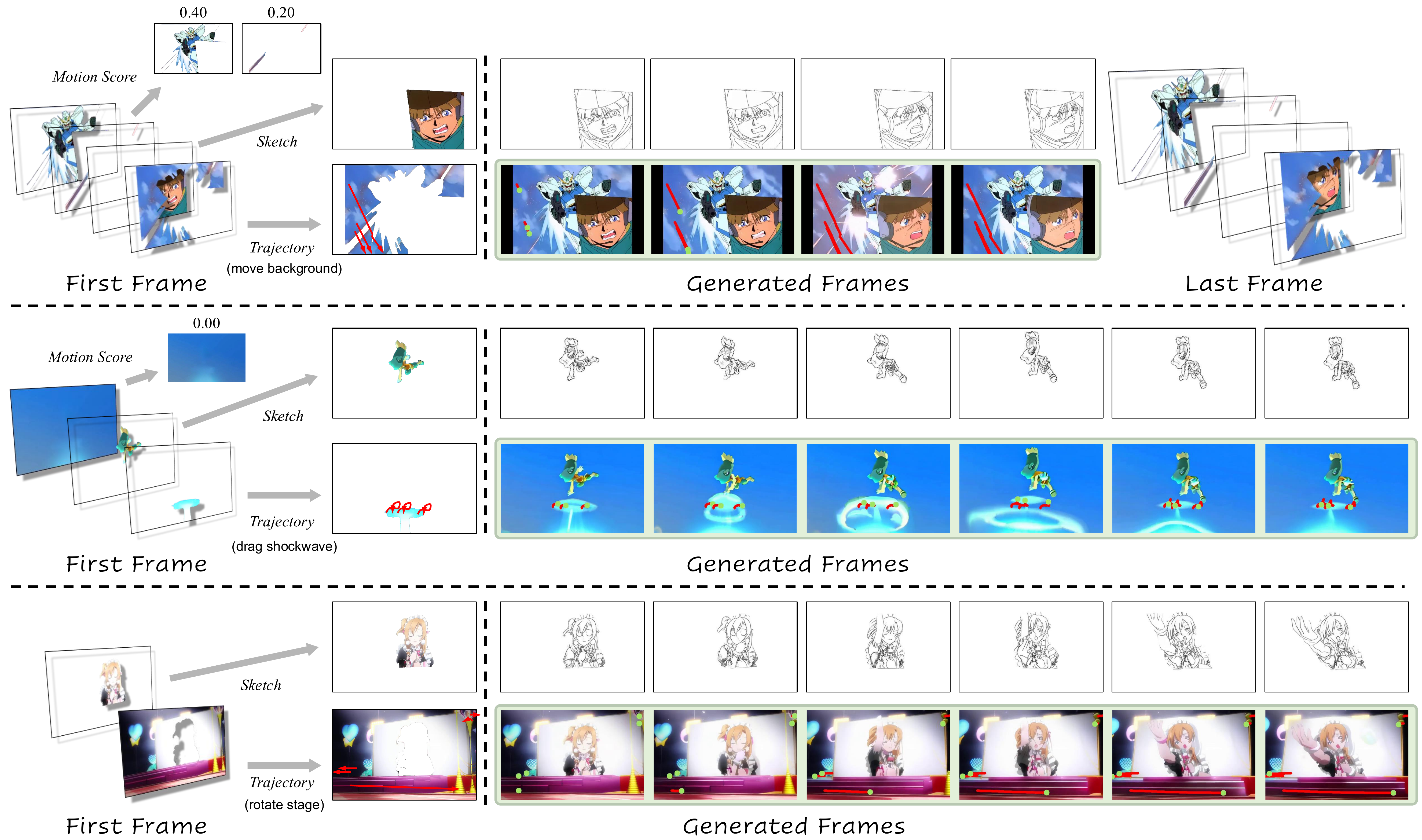}
    \vspace{-20pt}
    \caption{\textbf{Composite Control}. LayerAnimate provides multiple user-friendly control options at the layer level, leading to a composite control manner.
    We also provide the \textbf{corresponding videos} in the supplementary materials for clear illustration.}
    \vspace{-15pt}
    \label{fig:layer_control}
\end{figure*}
Our proposed LayerAnimate enables multiple heterogeneous control over animation layers.
Combining the multiple control signs leads to a composite manner of control, as illustrated in \cref{fig:layer_control}.
Taking the 4-layer sample as an example (the first row of \cref{fig:layer_control}), we employ sketch for the character layer to depict complex facial expressions while using trajectory movement for the sky and assigning different motion scores to mecha and light effects.
Ultimately, this approach enables the generation of animation clips with less cost than conventional frame-level sketching. Furthermore, other samples showcase effects such as the dragging of the luminous shockwave (the 2nd row) and the rotation of stages (the 3rd row).

\subsection{User Study}
To further evaluate the effectiveness of our method, we conduct a user study involving 20 participants who voted the best-generated videos among LayerAnimate and other competitors across six different tasks, as discussed in \cref{sec:comparison}.
As shown in \cref{fig:user_study}, our LayerAnimate exhibits superior performance.

\subsection{Ablation}
\label{sec:ablation}
\paragraph{Layer Capacity.} To investigate the impact of layer capacity $N$ settings on the performance, we test $N=1,2,4$ under I2V with motion scores condition.
As can be seen in \cref{tab:ablation}, increasing $N$ will improve the performance, demonstrating the superiority of our layer-level design.
In practice, using 4 layers is adequate in most animation cases, so we select $N=4$ as the default layer capacity.
\vspace{-12pt}
\paragraph{Motion Score.}
To demonstrate the effectiveness of layer-level motion information, here we progressively input motion information from binary motion state to specific motion score.
The binary motion state (i.e., w/ MA in \cref{tab:ablation}) means a certain layer is either static or dynamic.
Specific motion score provides more detailed information indicating the degree of movement, demonstrated by ``w/ MA \& scores'' in \cref{tab:ablation}.
As showcased in \cref{tab:ablation}, more motion information enables better generation quality in both I2V task and interpolation task.
\vspace{-12pt}
\paragraph{Trajectory Representation.}
For the representation of the trajectory, we integrate the commonly used Gaussian Heatmap and offset map forms in \cref{sec:layer_control_preparation}.
As demonstrated in \cref{tab:ablation}, our design results in a significant performance enhancement.
\input{table/ablation}
\section{Conclusion}
We propose \textit{LayerAnimate}, a layer-level control framework combining the traditional layer separation philosophy in animation production with video generation models.
LayerAnimate enables layer-level control over individual animation layers, allowing users to apply multiple controls to distinct layers.
To address the issue of scarce layer-level data, we design a data curation pipeline to automatically extract layer from animations.
Extensive experiments demonstrate its effectiveness and versatility.
This framework opens up new possibilities for layer-level animation applications and creative flexibility.

%% file: table/compare.tex
\begin{table}[ht]
\vspace{5pt}
\centering
\footnotesize
\begin{tabular}{c@{\hskip 5pt}c@{\hskip 5pt}l@{\hskip 3pt}c@{\hskip 5pt}c@{\hskip 5pt}c@{\hskip 5pt}c@{\hskip 5pt}c}

\toprule
\multicolumn{2}{c}{\textbf{Task}}& \textbf{Method} & \textbf{FVD$\downarrow$} & \textbf{FID$\downarrow$} & \textbf{LPIPS$\downarrow$} & \textbf{PSNR$\uparrow$} & \textbf{SSIM$\uparrow$} \\ 
\midrule
\multirow{10}{*}{\rotatebox{90}{I2V}}&\multirow{4}{*}{\rotatebox{90}{-}}
& SEINE~\cite{seine} &236.04&30.14&0.458&13.06&0.465\\ 
& & DynamiCrafter~\cite{dynamicrafter}&114.80&\textbf{14.36} &\textbf{0.354} &14.89 &0.554\\ 
&&CogVideoX~\cite{cogvideox}&170.31&19.77&0.355&14.35 &0.543 \\ 
&&LayerAnimate (ours) &\textbf{87.96} &14.66 &0.370 &\textbf{15.45} &\textbf{0.556} \\ 
\cmidrule{2-8}
&\multirow{3}{*}{\rotatebox{90}{Traj.}}
& DragAnything~\cite{draganything} &300.51 &26.10 &0.464 &14.00 &0.514  \\ 
&& Tora~\cite{tora} &190.61 &22.03 &0.376 &15.32 &0.525 \\ 
&& LayerAnimate (ours) &\textbf{72.04} &\textbf{12.55} &\textbf{0.281} &\textbf{17.46} &\textbf{0.634}\\ 
\cmidrule{2-8}
&\multirow{3}{*}{\rotatebox{90}{Sketch}}
& AniDoc~\cite{anidoc} &42.26 &12.16&0.131&21.39&0.792\\ 
&& LVCD~\cite{lvcd} &29.85 &7.01 &0.076 &\textbf{26.22} &\textbf{0.862}  \\ 
&& LayerAnimate (ours) &\textbf{26.64} &\textbf{5.92} &\textbf{0.075} &25.71 &0.858 \\ 
\midrule
\multirow{10}{*}{\rotatebox{90}{Interpolation}}&\multirow{5}{*}{\rotatebox{90}{-}} 
& SEINE~\cite{seine} &97.13 &11.96 &0.267 &18.19 &0.641\\ 
&& DynamiCrafter~\cite{dynamicrafter} &98.72 &13.03 &0.282 &17.95 &0.629\\ 
&& CogVideoX~\cite{cogvideox} &88.28 &9.93 &0.250 &19.39 &0.684 \\ 
&& ToonCrafter~\cite{tooncrafter} &74.63 &9.97 &0.244 &19.92 &0.668\\ 
&& LayerAnimate (ours) &\textbf{59.64} &\textbf{8.38} &\textbf{0.216} & \textbf{20.07}&\textbf{0.696} \\ 
\cmidrule{2-8}
&\multirow{2}{*}{\rotatebox{90}{Traj.}} &
Framer~\cite{framer} &62.17 &7.67 &\textbf{0.150} &22.48 &0.760 \\ 
&& LayerAnimate (ours)  &\textbf{44.69} &\textbf{6.94} &0.164 &\textbf{22.50} &\textbf{0.764}\\ 
\cmidrule{2-8}
&\multirow{2}{*}{\rotatebox{90}{\scalebox{0.9}{Sketch}}} &
ToonCrafter~\cite{tooncrafter} &66.26 &8.40 &0.128 &23.28 &0.794 \\ 
&& LayerAnimate (ours)  &\textbf{15.63}& \textbf{3.23}&\textbf{0.044} &\textbf{29.84} & \textbf{0.908}\\ 
\bottomrule
\end{tabular}
\caption{\textbf{Quantitative comparison with other state-of-the-art video generation models} across various tasks on our evaluation set. Traj.: Trajectory control.}
\vspace{-13pt}
\label{tab:comparison}
\end{table}

%% file: table/ablation.tex
\begin{table}[t]
\centering
\footnotesize
\begin{tabular}{c@{\hskip 4pt}l@{\hskip 5pt}|c@{\hskip 5pt}c@{\hskip 5pt}c@{\hskip 5pt}c@{\hskip 5pt}c}

\toprule
&Method & \textbf{FVD$\downarrow$} & \textbf{FID$\downarrow$} & \textbf{LPIPS $\downarrow$} & \textbf{PSNR$\uparrow$} & \textbf{SSIM$\uparrow$}\\ 
\midrule
\multirow{3}{*}{\rotatebox{90}{\scalebox{0.9}{Capacity}}}&I2V (N = 1)&87.88 &14.63 &0.376 &15.05 &0.546\\ 
&I2V (N = 2)&81.93 &14.15 &0.363 &15.39 &0.560\\ 
&I2V (N = 4)&\textbf{81.36} &\textbf{13.84} &\textbf{0.348} &\textbf{15.81} &\textbf{0.574}\\
\midrule
\multirow{6}{*}{\rotatebox{90}{\scalebox{0.9}{Motion Score}}}&I2V&87.96 &14.66 &0.370 &15.45 &0.556\\ 
&I2V w/ \textit{MA}&87.12 &14.44 &0.363 &15.64 &0.565\\ 
&I2V w/ \textit{MA} \& scores \dag &\textbf{81.36} &\textbf{13.84} &\textbf{0.348} &\textbf{15.81} &\textbf{0.574}\\ 
\cmidrule{2-7}
&Interp. &59.64 &8.38 &0.216 & 20.07&0.696\\ 
&Interp. w/ \textit{MA}& 59.79&8.31 &\textbf{0.214}&20.15&0.699\\ 
&Interp. w/ \textit{MA} \& scores &\textbf{53.42} &\textbf{8.03} &0.215 &\textbf{20.20} &\textbf{0.701}\\ 
\midrule
\multirow{3}{*}{\rotatebox{90}{\scalebox{0.85}{Represent.}}}&I2V w/ traj. (offset)&87.83 &12.74 &0.298 &16.94 &0.612\\ 
&I2V w/ traj. (heatmap)&80.57 &12.65 &\textbf{0.281} &\textbf{17.57} &\textbf{0.635}\\ 
&I2V w/ traj. (ours)&\textbf{72.04} &\textbf{12.55} &\textbf{0.281} &17.46 &0.634\\ 
\bottomrule
\end{tabular}
\vspace{-5pt}
\caption{\textbf{Ablation study on LayerAnimate.} \textit{MA}: Motion-based Assignment. Interp.: Interpolation. traj.: trajectory control. \dag: the same setting with ``I2V (N = 4)''.}
\vspace{-14pt}
\label{tab:ablation}
\end{table}

%% file: sec/X_suppl.tex
\clearpage

\setcounter{page}{1}
\maketitlesupplementary
\appendix

We provide \textcolor{blue}{videos on the project website\footnote{\url{https://layeranimate.github.io}}}.
These videos vividly present qualitative results and a novel application of multiple layer-level control for an enhanced view experience.
We recommend that readers watch these videos, as they provide a clearer and more intuitive understanding of this paper.





\section{Motion-based Hierarchical Merging}
\label{sec:mhm}
\begin{figure}[ht]
    \centering
    \includegraphics[width=0.8\linewidth]{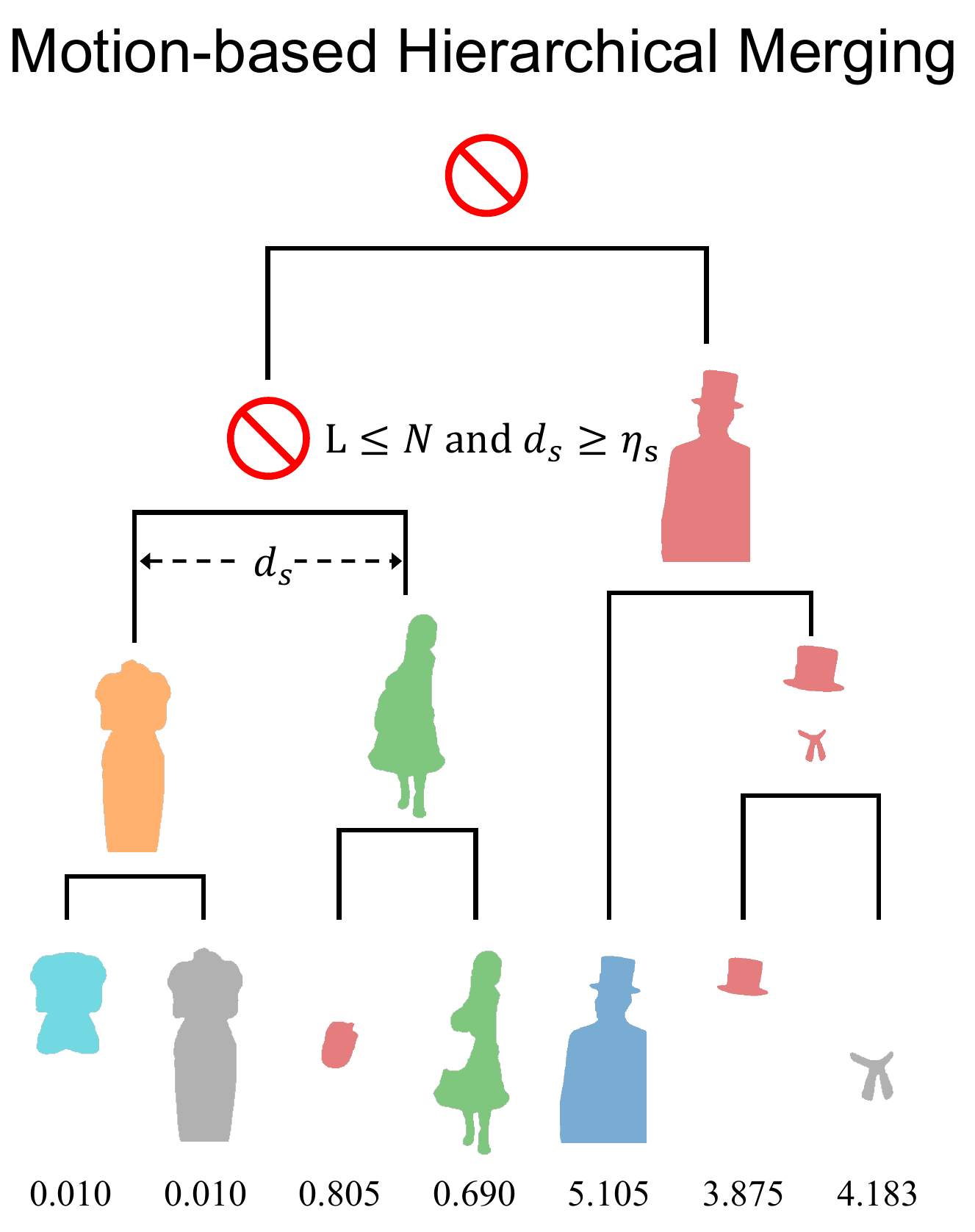}
    \caption{\textbf{Motion-based Hierarchical Merging.} Layers are merged from the bottom up until the layer count $L$ falls below the maximum layer capacity $N$ and the motion score difference $d_s$ exceeds the threshold $\eta_s$.}
    \label{fig:mhm}
\end{figure}

We showcase the illustration of \textit{Motion-based Hierarchical Merging (MHM)} in \cref{fig:mhm}.
MHM regards masklets as nodes and constructs a treemap using hierarchical clustering based on motion scores, merging layers with similar motion scores from the bottom up.
Considering the variability in layer numbers during production, we do not restrict a fixed number of layers.
Instead, we define the maximum layer capacity $N$, which is much less than the number of masklets, and a motion score merging threshold $\eta_s$.
Layers are merged from the bottom up until the count of layers falls below the capacity $N$ and the motion score difference exceeds the threshold $\eta_s$.

\section{Trajectory Control}
To fully elucidate the performance of trajectory control, we evaluate the Mean Squared Error (MSE) between the predicted animations and the ground truth object trajectories in \cref{tab:traj}.
\input{table/traj}
\section{Motion Score}
\begin{figure*}[ht]
    \centering
    \includegraphics[width=\linewidth]{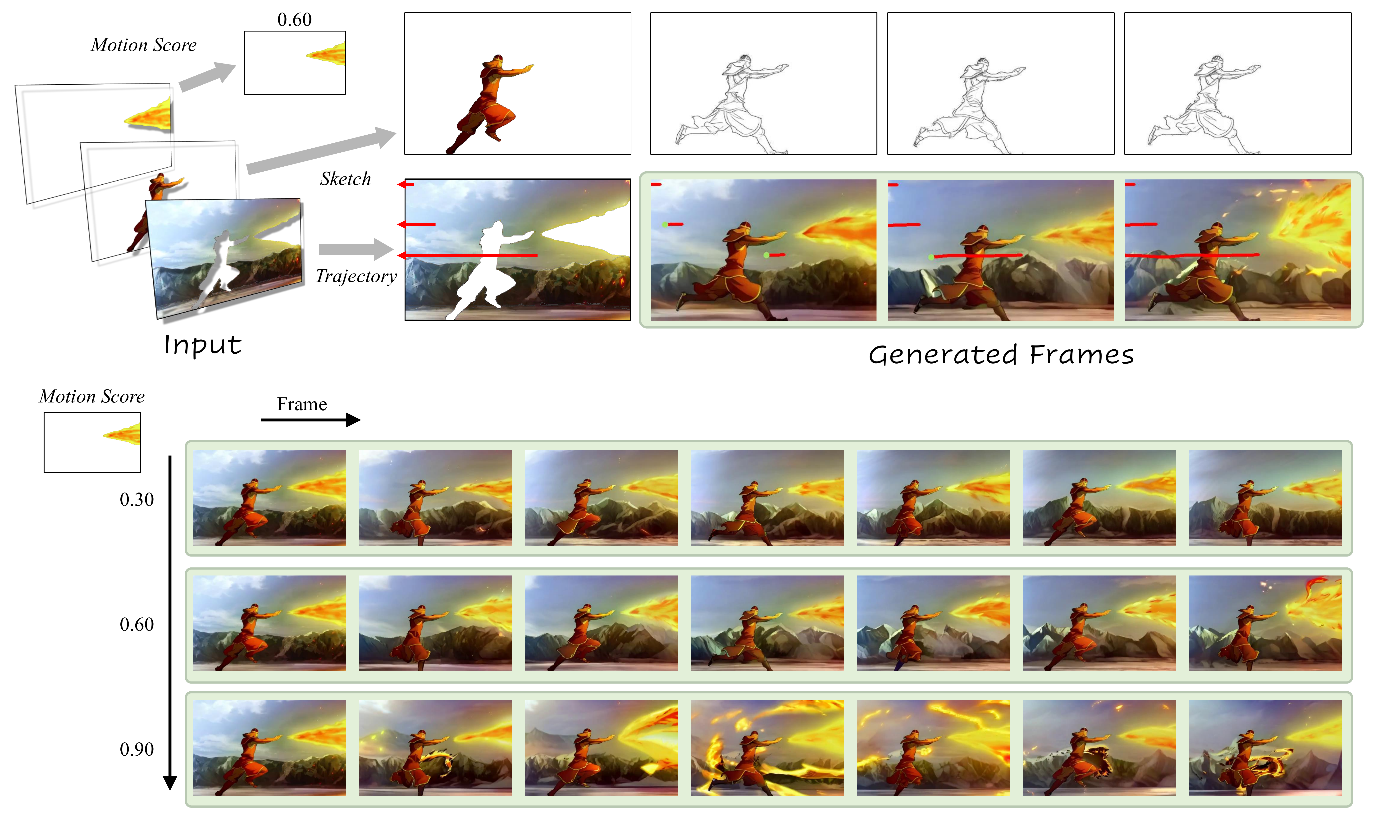}
    \caption{\textbf{Impact of different motion scores.}}
    \label{fig:supplement_score}
\end{figure*}

We vividly illustrate the impact of adjusting the motion score on generation in \cref{fig:supplement_score} using the sample from the teaser figure.
\section{More Qualitative Results}
\label{sec:supplement_qualitative}
\begin{figure*}[ht]
    \centering
    \includegraphics[width=\linewidth]{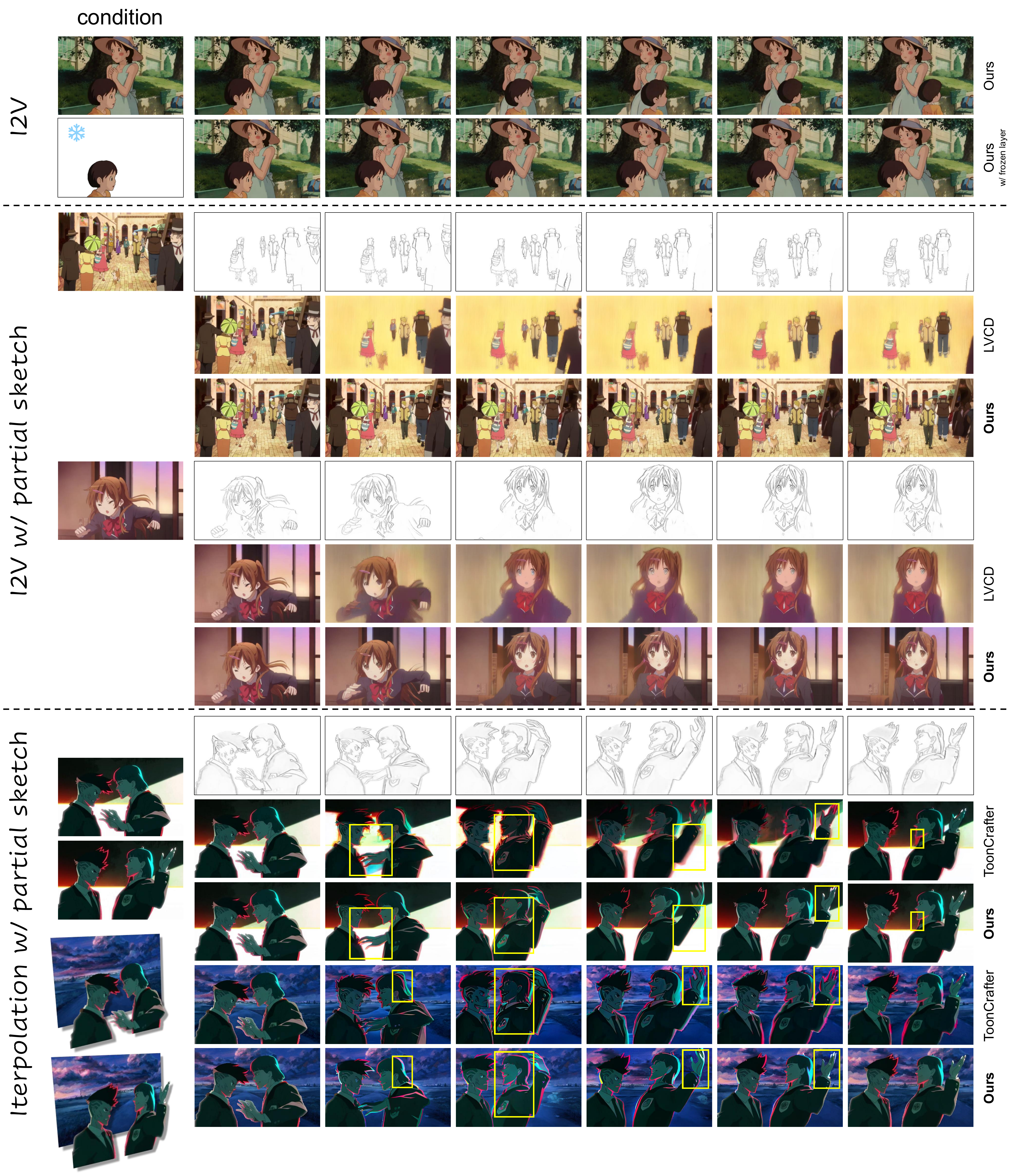}
    \caption{\textbf{Layer-level Application.} LayerAnimate provides innovative and user-friendly control options for animation, enabling users to freeze specific elements, animate characters with partial sketches, and switch dynamic backgrounds. The layer-level control over individual layers ensures that foreground layers remain consistent and nearly unaffected by background changes.}
    \label{fig:supplement_layer}
\end{figure*}

In this section, we provide additional application as illustrated in~\cref{fig:supplement_layer}.

\section{Limitations}
\label{sec:supplement_limitation}
While our approach introduces layer-level control tailored to animation, this concept presents opportunities for application in other data domains; for example, implementing layer-level control in real-world video generation based on depth information.

Additionally, we currently train the denoising UNet at a resolution of 512x320 with 16 frames, due to computational constraints.
In the future, we aim to enhance our framework by integrating more advanced video generation models, enabling animation generation at high-resolution and with longer durations.
    

%% file: table/traj.tex
\begin{table}[t]
\centering
\footnotesize
\begin{tabular}{l@{\hskip 5pt}|c@{\hskip 5pt}|c}

\toprule
Method & Task & \textbf{MSE$\downarrow$}\\ 
\midrule
DragAnything&I2V&1398.39\\ 
Tora&I2V &352.43\\ 
LayerAnimate (offset)&I2V&50.02\\
LayerAnimate (heatmap)&I2V&47.23\\
LayerAnimate (ours)&I2V&\textbf{46.58}\\
\midrule
Framer&Interpolation&44.91\\
LayerAnimate (ours)&Interpolation&\textbf{24.40}\\
\bottomrule
\end{tabular}
\caption{\textbf{Comparison of Trajectory Control Performance.}}
\label{tab:traj}
\end{table}